\documentclass[10pt,twocolumn,letterpaper]{article}

\usepackage{cvpr}              

\definecolor{cvprblue}{rgb}{0.21,0.49,0.74}
\usepackage[pagebackref,breaklinks,colorlinks,citecolor=cvprblue]{hyperref}

\def\paperID{65} 
\def\confName{CVPR}
\def\confYear{2024}

\title{REFA: Real-time Egocentric Facial Animations for Virtual Reality}

\author{Qiang Zhang \and Tong Xiao \and Haroun Habeeb \and Larissa Laich \and   Sofien Bouaziz \and Patrick Snape \and Wenjing Zhang \and Matthew Cioffi \and Peizhao Zhang \and Pavel Pidlypenskyi \and Winnie Lin \and Luming Ma \and Mengjiao Wang \and Kunpeng Li \and Chengjiang Long \and Steven Song \and Martin Prazak \and Alexander Sjoholm \and Ajinkya Deogade \and Jaebong Lee \and Julio Delgado Mangas \and ~ \\Reality Labs at Meta, Menlo Park, USA \and Amaury Aubel 
}

\begin{document}

\maketitle
\newcommand{\cllaich}[1]{}
\newcommand{\cmap}[1]{}
\newcommand{\SB}[1]{}
\newcommand{\hh}[1]{}
\newcommand{\qiang}[1]{}

\newcommand{\cpar}[1]{\textbf{#1}~}

\newcommand{\Appendix}[1]{Appendix~\ref{app:#1}}
\newcommand{\Section}[1]{Section~\ref{sec:#1}}
\newcommand{\Figure}[1]{Figure~\ref{fig:#1}}
\newcommand{\Equation}[1]{Equation~\ref{equ:#1}}

\newcommand{\Circle}[1]{\textcircled{\small{#1}}}
\newcommand{\inquote}[1]{``#1"}

\newcommand{\image}{\mathbf{I}}
\newcommand{\coefficients}{\mathbf{b}}
\newcommand{\translation}{\mathbf{t}}
\newcommand{\rotation}{\mathbf{R}}
\newcommand{\gaze}{\mathbf{g}}
\newcommand{\gazeright}{\mathbf{g}_\text{r}}
\newcommand{\gazeleft}{\mathbf{g}_\text{l}}

\newcommand{\real}{\mathbb{R}}
\newcommand{\rotationgroup}{\mathrm{SO(3)}}

\newcommand{\camera}{\mathcal{C}}
\newcommand{\visibility}{\mathcal{V}}
\newcommand{\region}{\mathcal{R}}
\newcommand{\normal}{\mathbf{n}}
\newcommand{\optical}{\mathbf{o}}

\newcommand{\range}{\mathcal{M}}
\newcommand{\expression}{\mathcal{E}}
\newcommand{\projection}{\mathcal{P}}
\newcommand{\keypoint}{\mathbf{K}}

\newcommand{\network}{\mathcal{N}}
\newcommand{\weight}{\boldsymbol{\theta}}
\newcommand{\texture}{\mathbf{T}}

\begin{abstract}
We present a novel system for real-time tracking of facial expressions using egocentric views captured from a set of infrared cameras embedded in a virtual reality (VR) headset. Our technology facilitates any user to accurately drive the facial expressions of virtual characters in a non-intrusive manner and without the need of a lengthy calibration step. At the core of our system is a distillation based approach to train a machine learning model on heterogeneous data and labels coming form multiple sources, \eg synthetic and real images. As part of our dataset, we collected 18k diverse subjects using a lightweight capture setup consisting of a mobile phone and a custom VR headset with extra cameras. To process this data, we developed a robust differentiable rendering pipeline enabling us to automatically extract facial expression labels. Our system opens up new avenues for communication and expression in virtual environments, with applications in video conferencing, gaming, entertainment, and remote collaboration.
\end{abstract}    

\section{Introduction}

Virtual reality (VR) transports users into simulated environments mimicking or enhancing real-world experiences. To achieve this, a head mounted display (HMD) presents three dimensional environments to users, along with other sensory feedback such as sound. Through these immersive experiences, users can interact with and explore computer-generated environments in a realistic manner. For example, users can tour a virtual museum, navigate through a digital city, or play a video games in VR. 

One particular important key feature in VR is the sense of \textit{social presence}, \ie people feeling that they are meaningfully interacting with others. 
This ability of feeling present with other people and form or deepen social connections is what makes VR truly engaging. However, to facilitate natural and intuitive social interactions, the development of accurate motion tracking technologies reproducing users' motions in real-time are required.

In particular, tracking facial motions is a key technology for social presence. This is achieved by capturing real-time video data of a person's face using cameras and then tracking specific features such as the mouth, nose, and eyes. By monitoring the movements of these features over time, face tracking detects and tracks facial expressions, such as smiles, frowns, and eyebrow raises. This signal can then be used to drive actions in a virtual environments, like \Circle{1} the challenge posed by occlusion of the user's face by the HMD, which makes it difficult to obtain an unobstructed capture of the face, \Circle{2} the complexity and cost of adding extra cameras to VR devices, and \Circle{3} the compute restrictions inherent to mobile platforms. 

\cpar{Contributions.} In this paper, we present an innovative system enabling accurate tracking of a user's facial expressions and movements using infrared (IR) cameras directly embedded within an HMD. Our contributions are \Circle{1} the placement of IR cameras and LEDs on an HMD through simulation, \Circle{2} an automated ground-truth generation pipeline allowing the collection of a large dataset using a lightweight capture process, \Circle{3} an iterative distillation framework allowing to train our machine learning (ML) model with heterogeneous and noisy labels acquired from different sources, and \Circle{4} an end-to-end system with auto-calibration and automated failure detection.

\section{Related Work}
The animation of digital characters through facial performance capture is a widely used technique within the computer graphics industry and has been a subject of ongoing research for many years. Pioneer works like Active Appearance Models (AAMs) \cite{edwards1998interpreting, cootes2001active, tzimiropoulos2013generic, matthews2004active} and 3D morphable models (3DMM) \cite{blanz1999morphable, knothe2011morphable, booth20163d} have been widely effective at registering faces in images by optimizing low-dimensional coefficients of linear subspaces for both shape and appearance. 

While AAM and 3DMM techniques effectively capture facial information within a linear model, the industry has predominantly embraced blendshapes subspaces \cite{lewis2014practice}. This preference arises from the semantic nature of blendshapes, enabling more meaningful and intuitive manipulation of facial expressions. This representation has been adopted with success by a large number of recent face tracking techniques \cite{weise2011realtime, bouaziz2013online, saito2016real}, including consumer products such as Apple's ARKit and Meta Spark. 

Beyond optimization based techniques, recent approaches \cite{richardson20163d, jourabloo2016large, zhu2016face, tuan2017regressing, guo2020towards} have leveraged deep learning techniques to regress low-dimensional coefficients from facial images. These methods typically train a convolutional neural network using large datasets to estimate the face shape, such as blendshape weights, from the input image. 

The accuracy limitations inherent to low-dimensional linear subspaces have led to the development of alternative approaches that directly generate detailed face shape and appearance as meshes and textures. To achieve high-quality results, some methods rely on a multi-camera rig \cite{dou2018multi, wu2019mvf} for capturing face shape and appearance. Other works focus on generating face shape and appearance from consumer devices, such as from RGB images \cite{ichim2015dynamic, Gecer_2019_CVPR, gecer2021fast, RingNet:CVPR:2019, DECA:Siggraph2021}, and from RGBD data \cite{zollhofer2011automatic, bouaziz2016modern, hifi3dface2021tencentailab}.

Another related topic involves generating face images from various inputs, such as text \cite{gecer2020tbgan, canfes2022latent3d}, audio \cite{richard2021meshtalk, prajwal2020lip, karras2017audio, VOCA2019}, or another face image (namely deepfake or faceswap) \cite{perov2020deepfacelab}. These techniques typically employ generative adversarial networks (GANs) to synthesize facial images. Additionally, some recent approaches utilize methods like Neural Radiance Fields (NeRF) \cite{athar2022rignerf} or stable diffusion \cite{kim2022diffface} for face image generation. These advancements have paved the way for generating face images from different modalities, enabling applications in text-to-face, audio-to-face, and image-to-face synthesis.


The most relevant work to ours is \cite{DBLP:journals/corr/abs-1808-00362, wei2019vr} which leverages an auto-encoder to drive avatars from cameras mounted on a VR headset. This model is trained using data captured from a large multi-camera rig, allowing for high-fidelity social interaction in virtual reality. Instead of training the model to decode both geometry and appearance, our model only decodes the geometry in blendshape format. This not only reduces computational costs but also allows for driving avatars of different appearances or styles, which means that 3rd party developers can use their own rigs with our model. Furthermore, we simplify the multi-camera rig used to train the model to a phone capture, which is available off the shelf. This enables us to collect a large and diverse dataset and train a model that can generalize to a broader population.


\section{Overview}
Our system utilizes 5 cameras that are integrated within a HMD (see \cref{sec:hardware}). These cameras capture infrared images at a resolution of 400x400 and a frame rate of 30 Hz. Our goal is accurately predict facial expressions from these camera images in real-time using an ML model. As a representation for the facial expression we choose 3D blendshape coeffcients which can be utilised to animate digital avatars. These 3D blendshape models (see \cref{sec:representation}) are a compact representation widely-used in previous work~\cite{blanz1999morphable} and in the industry.
To train our ML model we rely on three heterogeneous sources of data: \Circle{1} real, \Circle{2} synthetic, and \Circle{3} artist driven, each providing unique benefits (see \cref{sec:data}). To train our on-device model using these datasets gathered from different domains we employ an iterative distillation process (see \cref{sec:model}). To make our system robust to in the wild usage, we implement other key features as part of our end-to-end system design such as an online calibration step as well as a failure detection mechanism. Finally, we provide an extensive set of qualitative and quantitative evaluations to showcase the effectiveness of our approach (see \cref{sec:evaluation}).

\section{Hardware Design}\label{sec:hardware}


\begin{figure}[t]
     \centering
     \begin{minipage}[b]{1\linewidth}
  \centering
\includegraphics[width=1\linewidth]{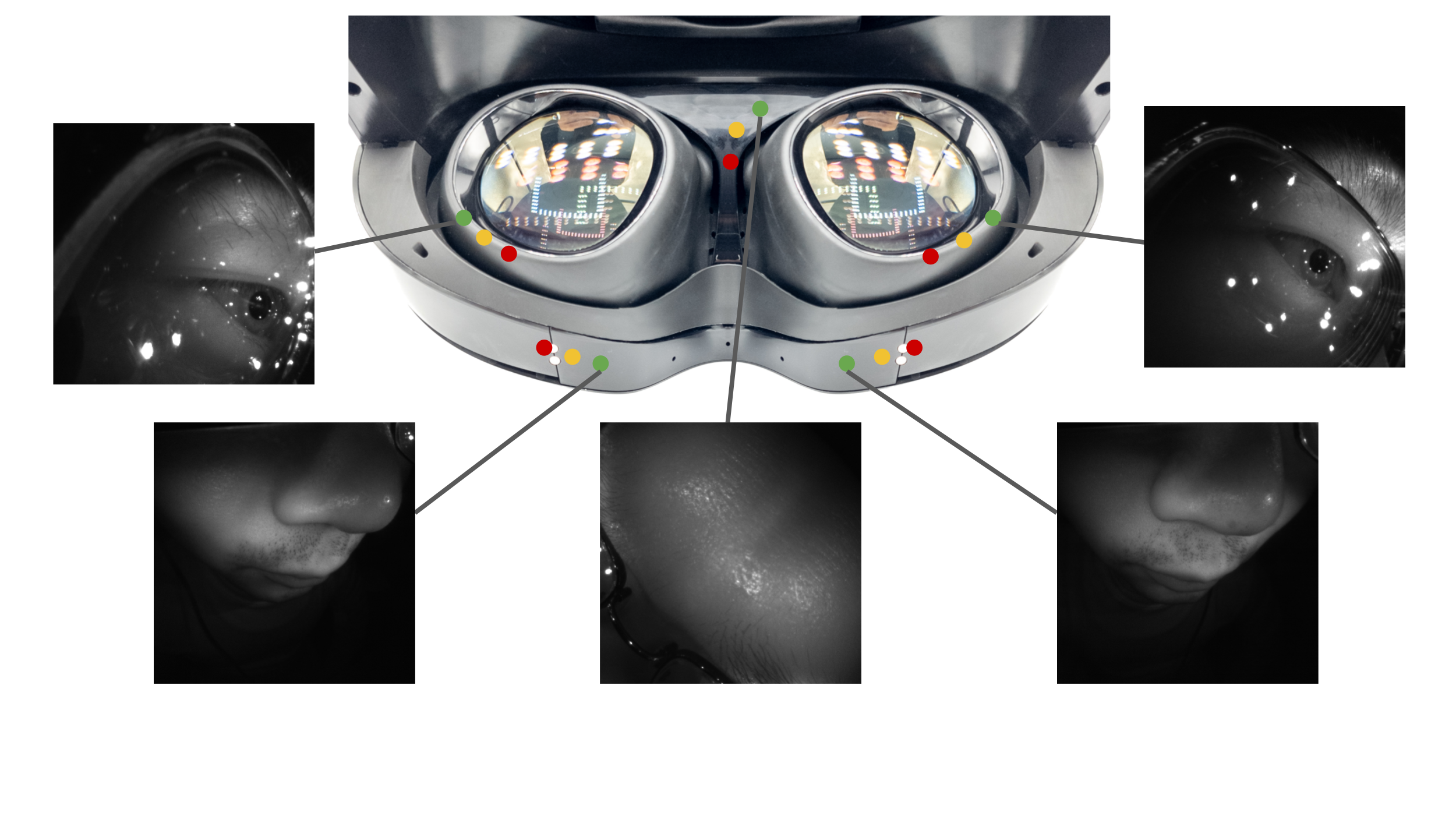}
     \end{minipage}
     \vfill
     \begin{minipage}[b]{1\linewidth}
\centering
\begin{tabular}{c|ccc}
\hline
\textbf{Eye}             & {\color{teal} Green}  & {\color{orange} Orange} & {\color{red} Red} \\
\hline
Visibility $\visibility$     & \textbf{0.508}& 0.494&0.473\\
Range of Motion $\range$   & \textbf{5.184}  & 3.782  & 4.203  \\
\hline
\hline
\textbf{Mouth}           & {\color{teal} Green}  & {\color{orange} Orange} & {\color{red} Red} \\
\hline
Visibility $\visibility$     &  \textbf{0.213}& 0.133& 0.104\\
Range of Motion $\range$ & \textbf{12.906} & 7.349  & 7.349  \\
\hline
\hline
\textbf{Glabella}        & {\color{teal} Green}  & {\color{orange} Orange} & {\color{red} Red} \\
\hline
Visibility $\visibility$     &0.268&0.351&\textbf{0.361} \\
Range of Motion $\range$ & \textbf{9.201}  & 8.143  & 6.274 \\
\hline
\end{tabular}
     \end{minipage}
 \caption{Our HMD is equipped with five face cameras, two for eye and eyebrow regions, two for mouth, and one for glabella. Note that we mirror the left eye and left mouth images. A multitude of camera configurations have been considered during the design of the HMD. Among these configurations, the one highlighted in green has been implemented, which has better visibility and range of motion metrics than the configurations highlighted in orange or red (Orange or red configurations seems to have better visibility in glabella, but they pose conflicts with users' glasses frames and the HMD's Inter-pupil distance adjustment mechanism).
 } 
 \label{fig:face_camera}
\end{figure}


We install 5 infrared cameras on our HMD, which are capable of capturing images at a resolution of $400\times400$ and a frame rate of 30 Hz. As shown in \cref{fig:face_camera}, two cameras are assigned to track eye and eyebrow movements, two are responsible for monitoring mouth movements, and the other one camera is designated to capture images of the glabella area, \ie, the region between the eyebrows. To cater to consumer use, the sensors are integrated within the hardware form factor, making it challenging to obtain a set of cameras with a clear view of the face. To resolve this issue, we used a PCA model generated from $800$ facial scans of $30$ expressions~\cite{vlasic2004multilinear}, which enabled us to assess different configurations for a wide range of facial structures and determine an adequate camera placement. To measure the quality of the different configurations, we measure the following metrics:

{\bf Visibility} $\visibility$: This metric quantifies the visibility of key facial regions $\region$. For a camera $\camera$, the visibility of a region is determined by the cosine of the angle between the surface normal $\normal$ of the face region and the optical axis of the camera $\optical$
\begin{equation}
    \visibility_\camera = \tfrac{1}{|\region|}\sum_{r \in \region} \normal_r^T\optical_{\camera}. 
\end{equation}
The weight ranges between 0 and 1, where a higher value indicates better visibility.

{\bf Range of Motion} $\range$: This metric assesses the ease with which a key set of facial expression $\expression$ can be captured by the camera $\camera$. It is determined by projecting the facial keypoints $\keypoint$ onto the 2D image space (denoted by $\projection(\cdot)$) for both neutral  and another facial expression $e \in \expression$, calculating the $\ell_2$ distance between these points, and then averaging across all the expressions
\begin{equation}
    \range_\camera = \tfrac{1}{|\expression|}\sum_{e \in \expression}\|\projection(\keypoint_e) - \projection(\keypoint_{neutral}) \|_F.
\end{equation}
The units of measurement are pixels, and a higher value indicates a greater range of motion, which is desirable.

We also introduced head pose variations when computing these metrics so as to make our camera placement robust to donning preferences. By simulating and incorporating these variations, we aimed to ensure that the sensors perform reliably and accurately regardless of how the headset is donned by the user. \Cref{fig:face_camera} shows a few considered camera locations and their corresponding metrics.



\begin{figure}[t]
     \centering
     \begin{minipage}[b]{1\linewidth}
  \centering
  \includegraphics[width=1\linewidth]{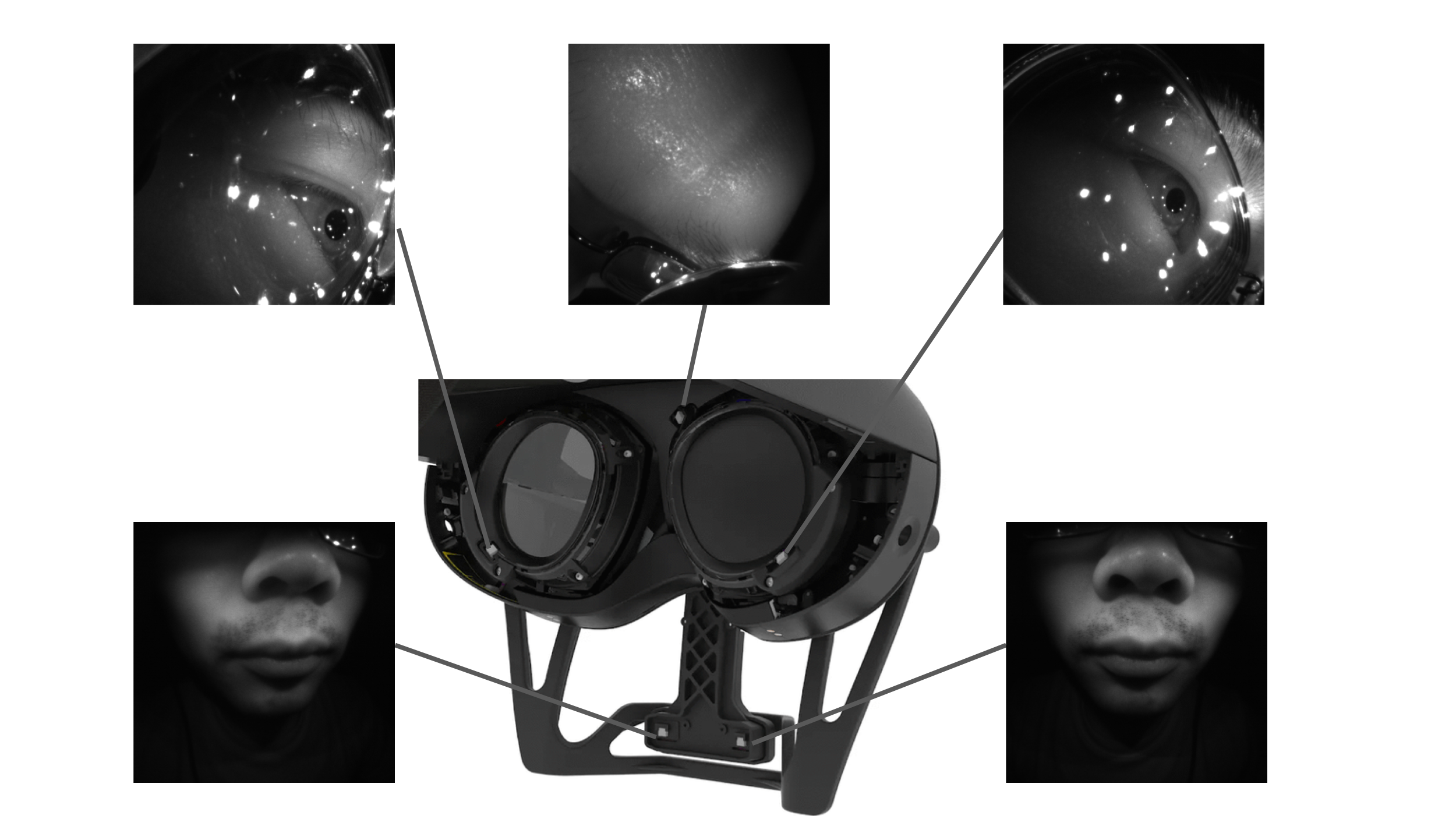}
     \end{minipage}
     \vfill
     \begin{minipage}[b]{1\linewidth}
 \includegraphics[width=1\linewidth]{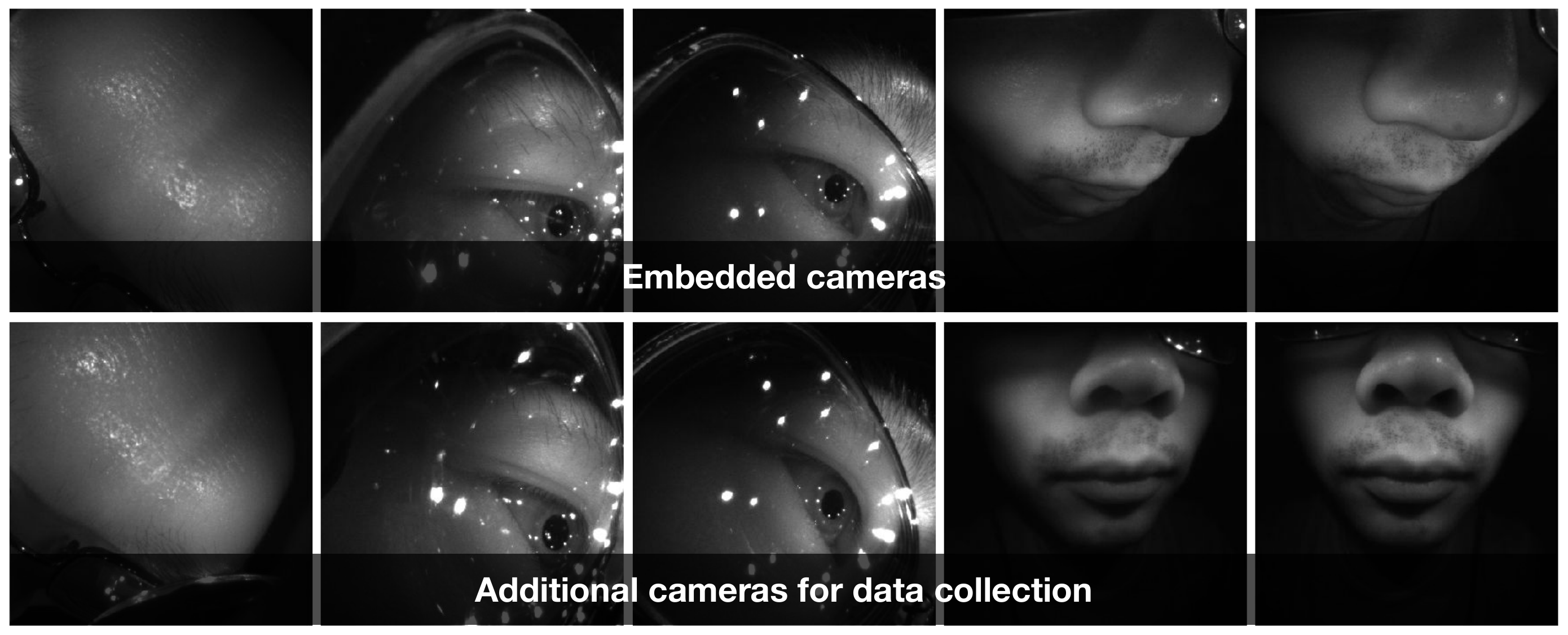}
     \end{minipage}
 \caption{Our data collection HMD is equipped with additional five cameras, offering better visibility of the face than the embedded cameras. This camera setup allows us to improve the quality of the generated pseudo ground truth.}
 \label{fig:arcata_dd}
\end{figure}

\section{3D Face Representation} \label{sec:representation}

A central component of our system is a blendshape model~\cite{blanz1999morphable} that provides a low-dimensional representation of the user’s expression space based on Ekman’s Facial Action Coding System (FACS)~\cite{ekman1978facial}. Our blendshape model contains $53$ bases, which correspond to 3D meshes that can be combined linearly to produce new facial expressions. To combine these bases, we use a weight vector of blendshape coefficients $\coefficients \in \real^{53}$, where each entry falls within the range of $\left[0.0,1.0\right]$. A weight of 0.0 indicates an inactive expression, while a weight of 1.0 signifies full activation and is the maximum movement a person can perform. Our model also incorporates eye gaze vectors $\gazeleft \in \real^2$ and $\gazeright \in \real^2$ for the left and right eyes, respectively. These vectors are also used to device an additional set of $8$ eye following blendshapes. Additionally, the face's rigid motion is parameterized using a translation vector $\translation \in \real^3$ and a rotation matrix $\rotation \in \rotationgroup$.


\begin{figure}[t!]
    \centering
    \includegraphics[width=\linewidth]{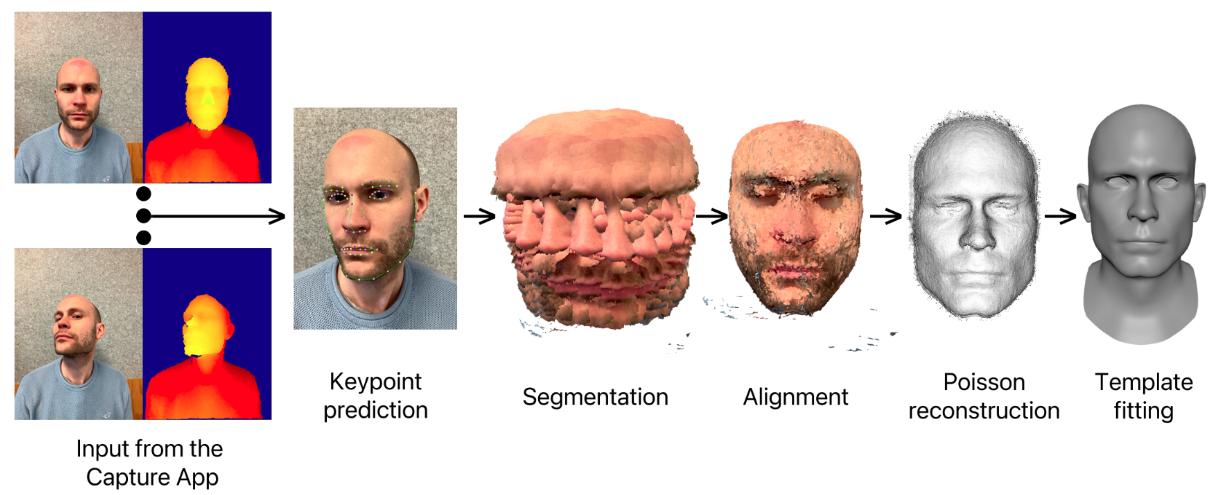}
    \caption{
Our expression fitting pipeline takes RGBD frames as input and proceeds through a series of steps to generate a fitted mesh as output. We use this process to fit a set of facial expressions, from which we then create a subject-specific rig using example-based facial rigging \cite{li2010example}.}
    \label{fig:mobile_genesis}
\end{figure}

\section{Data Generation} \label{sec:data}
To train our system we require a large dataset of camera frames labeled with blendshape coefficients. We collect a real-world dataset of 18k subjects providing 3 trillion frames. To annotate this extremely large number of frames, we develop an automated self-supervised approach based on a differentiable renderer (see \cref{sec:realdata}). Further, a set of technical artists annotated a key set of frames manually to provide semantic labels (see \cref{sec:artistdata}). Finally, we complement our real-world dataset using synthetic data providing exact labels for challenging or long-tail cases (such as facial hair, glasses, donning variation, \etc, see \cref{sec:syntheticdata}).

\subsection{Real Data} \label{sec:realdata}
\subsubsection{Capture Process} To collect a large dataset of a diverse set of subjects we develop a lightweight capture setup based on a mobile phone containing a depth sensor (iPhone 12) as well as a modified HMD. Capturing accurate views of the face presents a fundamental challenge with our embedded camera setup, primarily due to the close placement of the cameras, resulting in occlusions that obstruct the field of view. To alleviate this issues, we developed a data collection HMD equipped with an additional five ground truth cameras with better visibility including two boom cameras capturing the lower face from a frontal angle, two eye ground truth cameras, and one glabella ground truth camera (see \cref{fig:arcata_dd}). The inclusion of these additional ground truth cameras, in conjunction to the five embedded ones, offers alternative view angles significantly enhancing the visibility of the face (see \cref{fig:arcata_dd}).






\noindent \textbf{Mobile phone capture.} We utilize the mobile phone to gather a diverse set of 60 individual facial expression scans. Subjects are instructed to hold specific facial expressions while making slight head movements in front of the mobile phone, enabling us to collect RGBD frames from multiple angles.

\noindent \textbf{HMD capture.} Using our modified HMD, we request subjects to engage in a series of facial motions. This allows us to capture approximately 40 minutes of motion sequences encompassing a diverse range of expressions and speech sequences.



\subsubsection{Generating subject-specific blendshape rig from mobile phone data} \label{sec:mg}

The process of generating facial blendshapes involves a multi-step optimization problem, performed individually for each subject. For each captured expressions, we first predict 100 facial keypoints~\cite{newell2016stacked} per frame, and extract a segmentation mask for the face~\cite{he2017mask}. We then align the RGBD frames using rigid ICP~\cite{bouaziz2016modern} and merge the result using Poisson reconstruction~\cite{kazhdan2006poisson}, obtaining a reasonable integrated mesh (see \cref{fig:mobile_genesis}). In order to maintain topological consistency across different subjects' meshes, we employ a PCA model~\cite{vlasic2004multilinear} that is fitted in conjunction with head pose estimation. This is followed by a refinement step using a Laplacian non-rigid deformation technique~\cite{bouaziz2016modern}. Finally, we compute the personalized blendshape rig using example-based facial rigging \cite{li2010example}.

\subsubsection{Estimating blendshape coefficients from HMD images} \label{sec:mr}

\begin{figure}[t!]
    \centering
    \includegraphics[width=\linewidth]{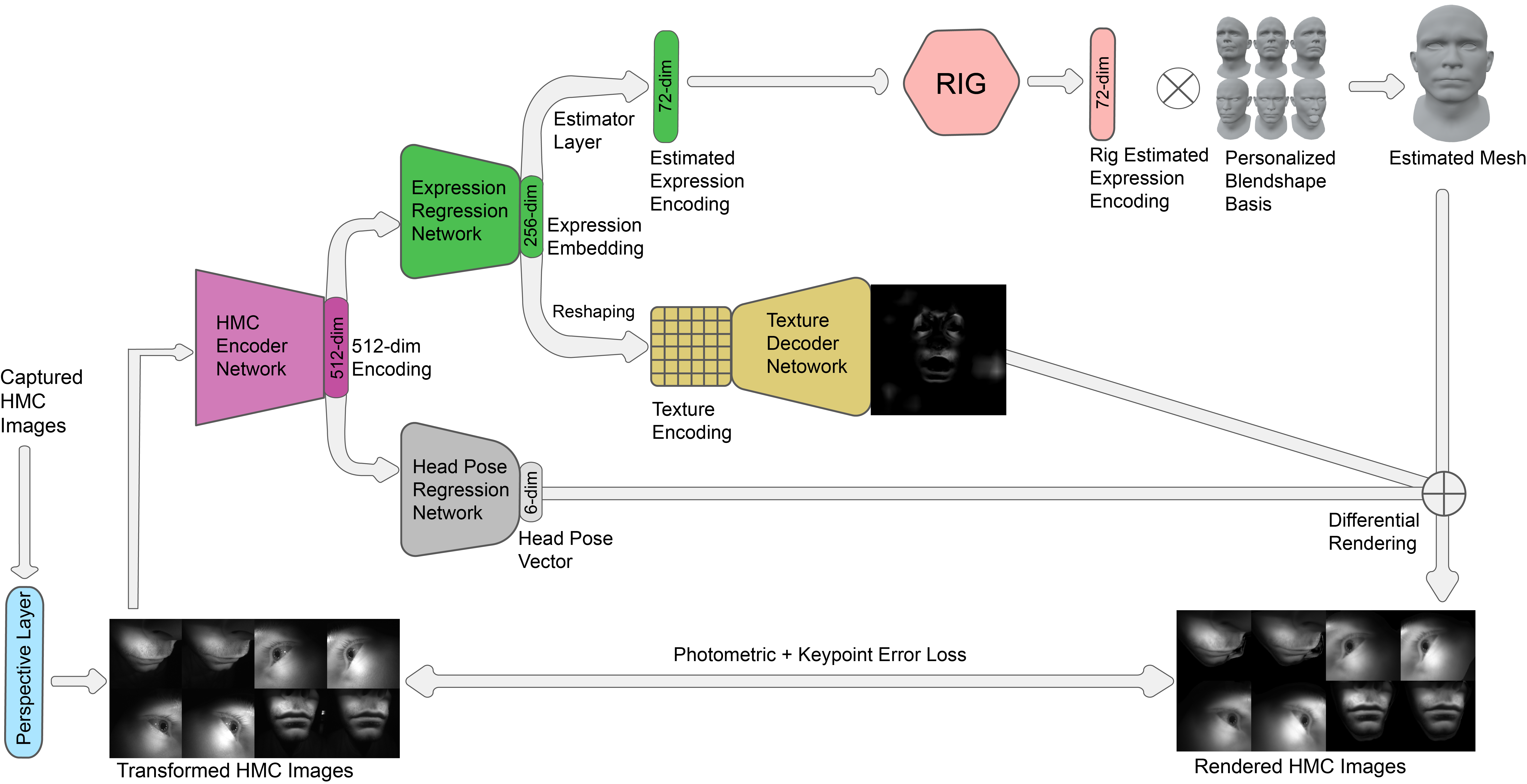}
    \caption{System diagram of estimating blendshape coefficients from HMD images, based on the subject-specific blendshape rig.}
    \label{fig:mobile_rosetta}
\end{figure}

Given a subject-specific blendshape rig, we solve a ``self-supervised'' learning problem per subject to establish correspondences between input HMD images and output blendshape coefficients. We parameterize our problem with a Convolutional Neural Network (CNN) $\network_{{\weight}} : (\image)\rightarrow (\coefficients, \rotation, \translation, \texture)$, with the goal of predicting per frame blendshape coefficients $\coefficients$, head pose $(\rotation, \translation)$ and texture $\texture$ from the input images $\image$, which has been proven beneficial~\cite{10.1145/3306346.3323030}. By leveraging the blendshape coefficients and head pose information, we are able to reconstruct the mesh. This reconstructed mesh, along with the corresponding texture, can then be rasterized to reconstruct the input images. Our rasterizer is differentiable~\cite{10.1145/3306346.3323030} and we optimize for the network's weights $\weight$ using Adam~\cite{kingma2014adam}. We use three losses, \Circle{1} a keypoint $\ell_2$ reprojection error, \Circle{2} the $\ell_2$ pixel differences between the input and reconstructed views, and \Circle{3} a $\ell_1$ sparsity regularization of the blendshape coefficients. See \cref{fig:mobile_rosetta} for a detailed architecture of our approach. 

The source of our blendshape bases are artist-provided sculpted meshes, with each mesh corresponding to a FACS shape \cite{ekman1978facial}. These bases do not form a set of independent vectors and in some cases the same mesh can be generated using different blendshape coefficients. The $\ell_1$ sparsity regularization used during the network optimization helps to regularize this problem but does not fully solve it. To further improve, we add a set of semantic ``rig constraints'' during the optimization.





\begin{figure}[t!]
    \centering
    \includegraphics[width=\linewidth]{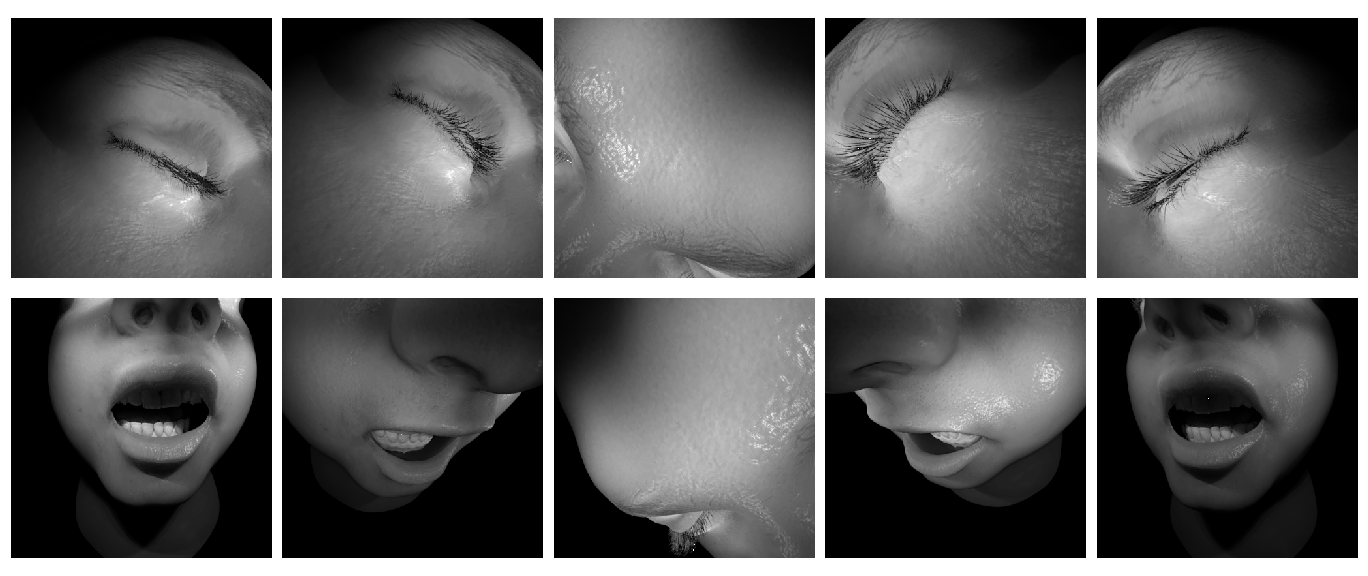}
    \caption{Example of a synthetic frame used to train our ML model}
    \label{fig:synthetic_example}
\end{figure}

\subsection{Art Priors} \label{sec:artistdata}
\begin{figure}
    \centering
    \includegraphics[width=\linewidth]{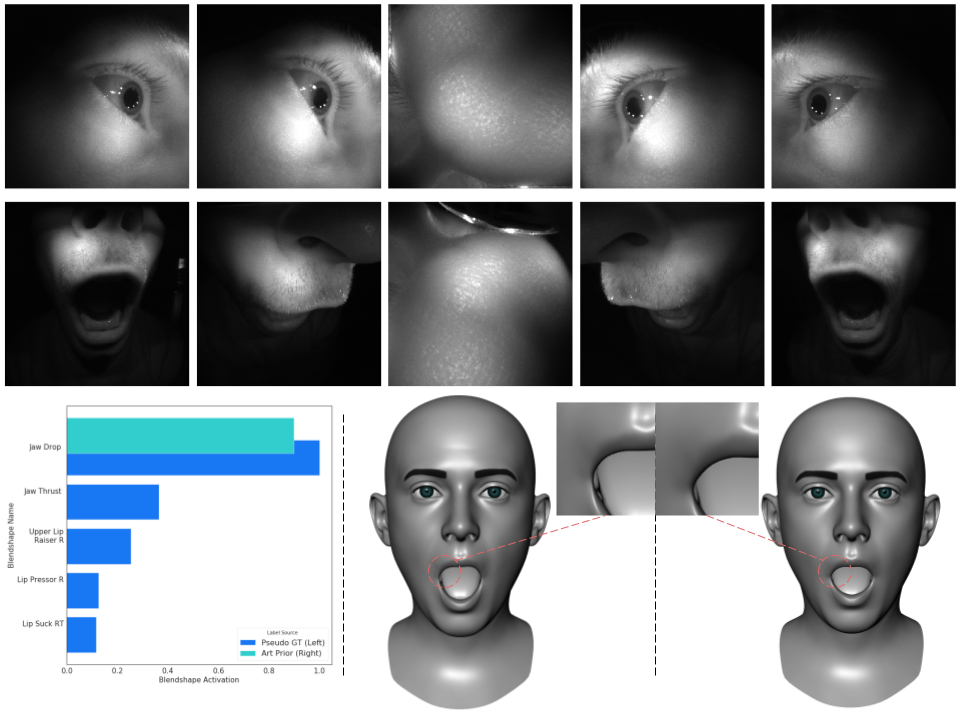}
    \caption{An example frame for the peak ``Jaw Drop Max'' expression. The left avatar is the pseudo ground truth generated based on the method described in \cref{sec:mr}. The right avatar represents the art priors. The histogram in the bottom right displays top-5 activated blendshapes for this frame. Note that the art priors are sparse compared with the pseudo ground truth. While the art priors are not perfectly accurate, they can be semantically meaningful for peak expression frames.}
    \label{fig:patrick_artgt}
\end{figure}

In our data collection, we include multiple segments where participants begin from a neutral pose, performing straightforward expressions by following a reference photo/video prompt that helps the person to mimic, and then revert to the neutral position. For each of these peak expressions, we ask FACS experts and skilled artists to define a set of expected blendshape coefficient activations, as demonstrated in \cref{fig:patrick_artgt}. While not perfectly accurate, these labels can be considered as art priors, which are used later in training (\cref{subsec:training-framework}) and evaluating (\cref{sec:evaluation}) the on-device ML models.


\subsection{Synthetic Data} \label{sec:syntheticdata}
We generate a large synthetic dataset of roughly 25 million frames from 800 identities collected with a multi-view capture system and rigged with the methodology described in \cref{sec:mg} (see \cref{fig:synthetic_example}). We retarget animation sequences generated by artists and sample $\sim$30000 frames per identity. We increase the variation of the artist generated animation sequences by also including segments that were rare in our real dataset. To improve realism and increase diversity, we augment the data with facial hair and glasses. The hair generation process involved procedurally growing hair and fitting it to the scalp of the rigged model. As for glasses, we employed 3D assets created from a set of frontal scans to generate a wide variety of glasses designs, which were then fitted to the rigged model. Although a visual domain gap persists between real and synthetic data, this synthetic dataset offers perfect labels that can be effectively utilized to improve the accuracy of our ML model.


\section{On-Device ML Model} \label{sec:model}

\subsection{Architecture}
\begin{figure}[t!]
    \centering
    \includegraphics[width=\linewidth]{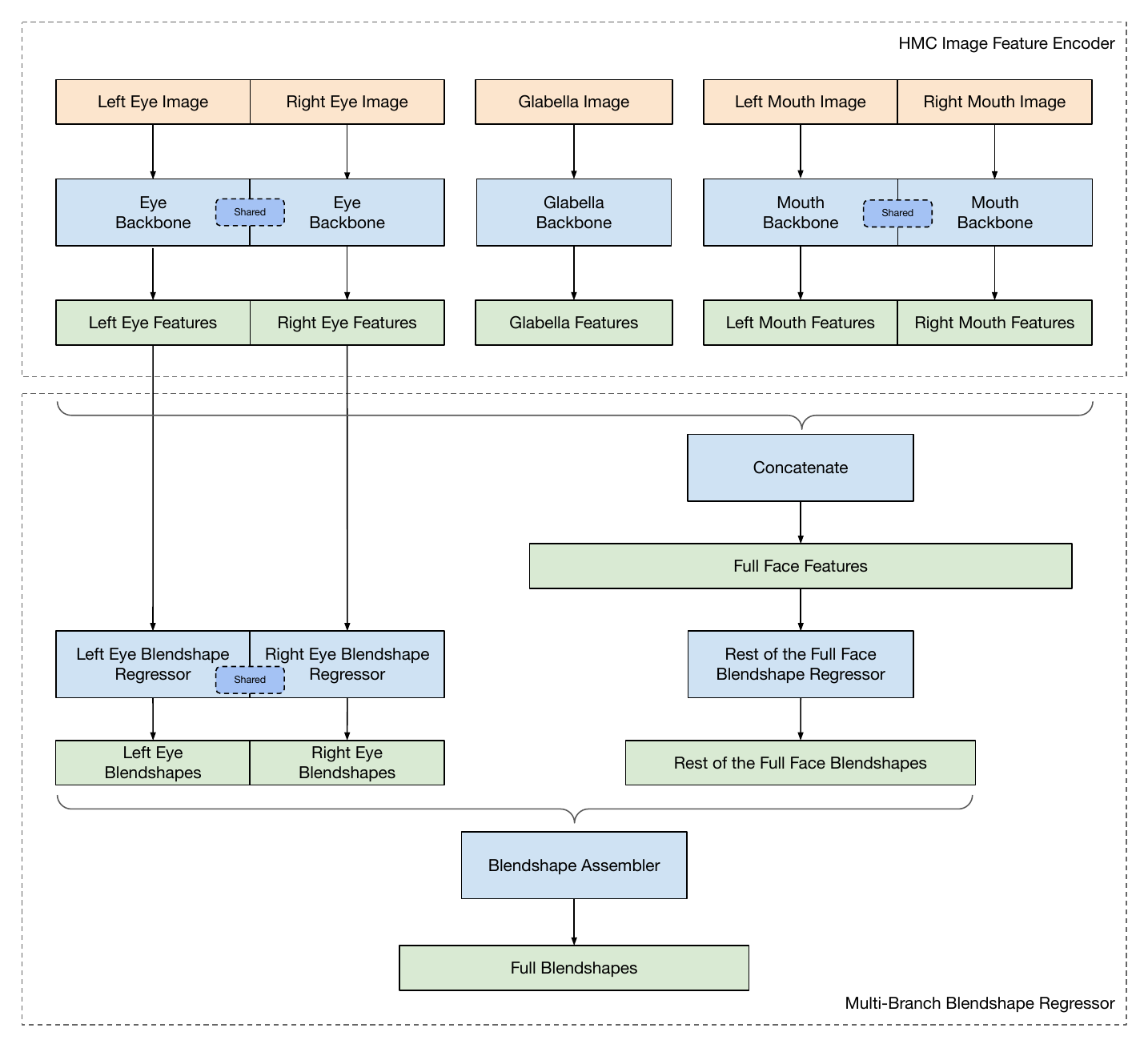}
    \caption{Architecture of the on-device ML model}
    \label{fig:on-device-model}
\end{figure}
Using the dataset described in \cref{sec:data}, we train a CNN that is specifically designed for efficient on-device inference. As illustrated in \cref{fig:on-device-model}, the model consists of a headset-mounted camera (HMC) image feature encoder and a multi-branch blendshape regressor.

\noindent\textbf{HMC image feature encoder.} We employ a ResNet-like~\cite{he2016deep} CNN backbone to extract features from each of the HMC images. We flip the images acquired from the left eye and the left mouth cameras allowing us to share the parameters between the left and right, eye and mouth backbone models, respectively. The input images are resized to $224\times 224$. The $512$-channel output feature maps are of resolution $7 \times 7$, which are then averaged-pool into a feature vector of $512$ dimensions for each image.

\noindent\textbf{Multi-branch blendshape regressor.} We estimate the final blendshape coefficients by assembling the per-image features from the different branches. The left eye features are fed into a multi-layer perceptron (MLP) to estimate the left eye blendshapes. Similarly for the right eye, which shares the same MLP parameters as the left eye. Finally, features from all the cameras are concatenated to estimate the blendshapes for the rest of the face. In our dataset the left and right eyes are often open or closed simultaneously leading the ML model to learn such predominant correlation if simply concatenating all the features and estimating all blendshape coefficients at once. Our network design enables to better capture rare asymmetric upper face expressions (such as winking) by explicitly breaking correlations in the network architecture.

\subsection{Training Framework} \label{subsec:training-framework}

We train our ML model end-to-end with an $\ell_1$ loss between the estimated blendshapes and the labels using the Adam optimizer~\cite{kingma2014adam}. However to efficiently use the synthetic dataset we need to modify our training framework to accommodate for the domain gaps between the different modalities. 

\begin{figure}
    \centering
    \includegraphics[width=\linewidth]{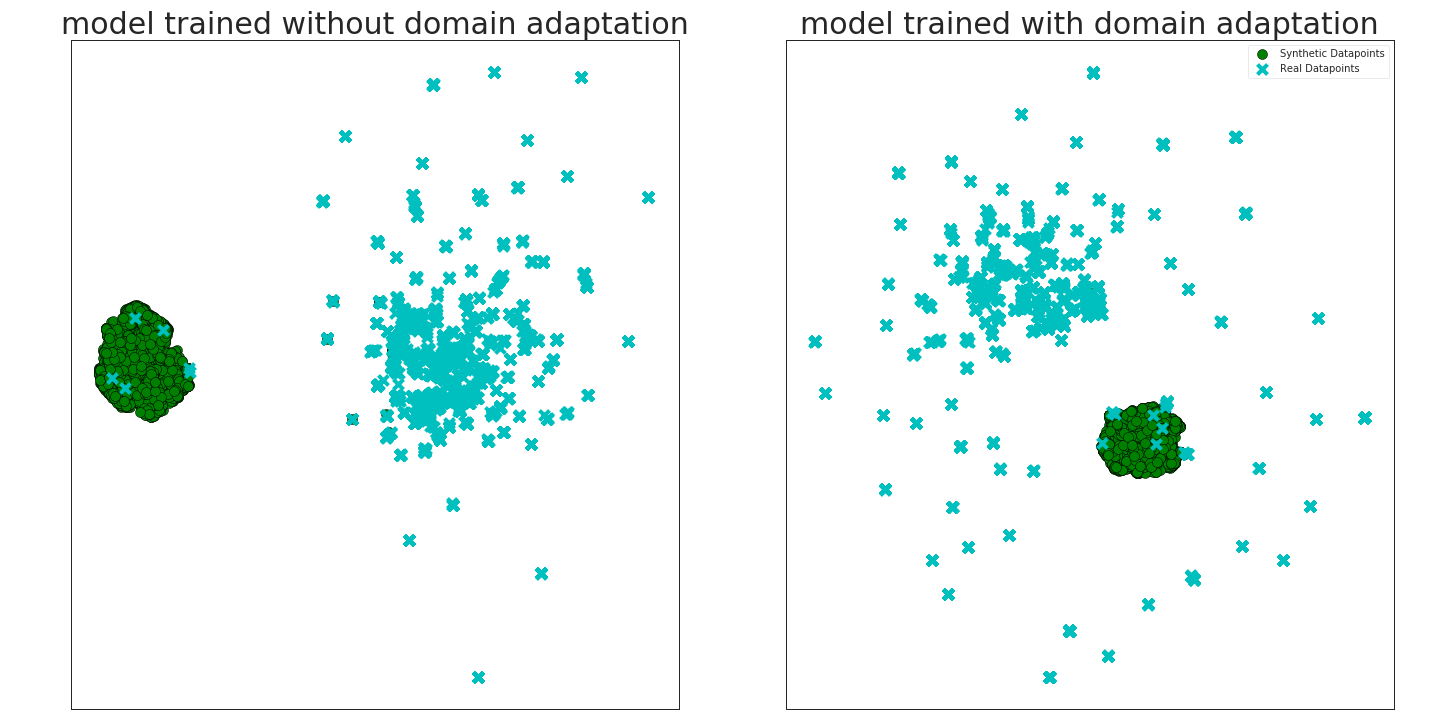}
    \caption{Visualize the ML model's feature space (with UMAP~\cite{SMG2020}) for real data (blue crosses) and synthetic data (green dots). Left: model trained \textbf{without} domain adaption. Right: model trained \textbf{with} domain adaptation. Domain adaptation better aligns the features between the real and synthetic data domains.}
    \vspace{-3ex}
    \label{fig:synthetic_z_umap}
\end{figure}

\noindent\textbf{Domain Adaptation.}
As noted in \cref{sec:syntheticdata}, although the synthetic data looks fairly close to the real data, a significant domain gap still exists. The distinction between real and synthetic data becomes evident when observing the disparity in feature distribution, as demonstrated in \cref{fig:synthetic_z_umap}.

This existing domain gap poses a challenge for the ML model to effectively leverage the knowledge present in both the synthetic and real datasets. To address this challenge, we follow the gradient reversal technique proposed in~\cite{ganin2016domain} which performs feature-level domain adaptation during the ML model training. Specifically, we first build a mini-batch with equal numbers of real and synthetic data frames. Then, we train an additional domain discriminator with binary cross entropy loss based on the image features. Finally, the gradient from the discriminator is reversed before backpropagating to the feature encoders. As demonstrated in \cref{fig:synthetic_z_umap}, this procedure better aligns the features between the two domains, letting the ML model further benefit from the samples additionally provided by the synthetic data.

\noindent\textbf{Iterative Distillation.} During our experiments, we discovered that naively training the ML model on real data labels resulted in expressions that appeared ``muted''. For instance, when a person fully raised their eyebrows, the avatar's eyebrows would only exhibit a slight movement. This discrepancy arose because real data labels have been generated automatically on a per-subject basis using a self-supervised learning technique (\cref{sec:mr}). As a consequence, these labels contain inherent noise and outliers, which forces the network to produce some amount of averaging to fit the noisy distribution of blendshape coefficients.

Inspired by student-teacher approaches that have been used to learn from noisy labels~\cite{xie2020self,hinton2015distilling}, we propose an iterative training algorithm to address this challenge (see \cref{alg:iterative_distillation}). As part of the iterative training process, we also incorporate the Art Priors (\cref{sec:artistdata}) to refine and improve the network's ability to generate more accurate and meaningful results.

Our process starts with the initial real data $R_0$ and synthetic data $S$. Then, we repeat the distillation process for $T$ iterations as follows:

\begin{algorithm}
\caption{Iterative distillation algorithm}\label{alg:iterative_distillation}
\DontPrintSemicolon
\SetKwFunction{Train}{Train}
\SetKwFunction{PostProcess}{PostProcess}
\SetKwFunction{Select}{Select}
\SetKwFunction{EnsembleInference}{EnsembleInference}
\KwData{Initial pseudo-labeled real data $R_0$, synthetic data $S$}
\KwResult{Final on-device ML model $m$}
\For{$t\leftarrow 1$ \KwTo $T$}{
    Model Pool $M\leftarrow$ \Train{$R_{t-1}$, $S$}\;
    $M\leftarrow M \cup$ \Train{\PostProcess{$R_{t-1}$}, $S$}\;
    Best Performed Models $M^*\leftarrow \Select{M}$\;
    $R_t\leftarrow$ \EnsembleInference{$M^*$, $R_{t-1}$}\;
}
$m\leftarrow$ \Train{\PostProcess{$R_T$}, $S$}
\end{algorithm}

In each iteration, we first train a set of models with the pseudo-labeled real data and the synthetic data, with different random seeds. Second, we post-process the pseudo labels with Range-of-Motion calibration and temporal smoothing, and train another set of models. Third, we select several best performed models based on quantitative metrics (\cref{section:metrics}). Last, we infer the best performed models over the real data and ensemble the inference results to be the new pseudo labels.

Empirically, the quantitative metrics often stop to improve after 5 or 6 rounds. We thus train the final on-device ML model with the last post-processed pseudo-labeled real data and the synthetic data.

This algorithm can be viewed as using the ML models to iteratively refine the initial pseudo labels toward the ``true'' ground truth. Since the true ground truth is unknown, the quantitative metrics serve as a proxy to measure how close we are.

\section{Evaluation}\label{sec:evaluation}
To thoroughly and comprehensively evaluate the effect of our face tracking solution to the \textbf{end-user experience}, we propose and develop three tiers of evaluation approaches.

\noindent\textbf{Quantitative Metrics.} Automatically compute several heuristic blendshape-based metrics that we found correlated with the user experience.

\noindent\textbf{Qualitative Evaluation (QE).} Render the tracking results as avatars alongside the corresponding camera images. Send to human annotators to rate whether the tracking results match the expressions performed.

\noindent\textbf{User Experience Research (UXR).} Build VR Apps for 1-on-1 conversation and small group meetings. Integrate our face tracking solution to the Apps and allow it to be turned on/off. Recruit a group of diverse users to try out the VR Apps and thoroughly evaluate the experience through questionnaires.

Based on the quantitative metrics, we further analyze how the iterative distillation (\cref{sec:ablation_iterative}) and the training dataset size impact on the accuracy of the ML model. Lastly, we discuss the limitations of our system.

\subsection{Quantitative Metrics}\label{section:metrics}
Conventional metrics, such as comparing the $\ell_1$/$\ell_2$ distance between the estimated and ground truth (GT) blendshape coefficients or mesh vertices, are less effective in our case. Because 1) we only have the pseudo GT, not necessarily the true GT, and 2) the conventional metrics do not correlate well with the end-user experience, \eg, when the user is gently smiling, the tracking results of larger smiling and gently frowning could lead to similar mesh errors, but they mean very differently to the user.

To address the challenges, we propose a set of heuristic metrics that are comprehensive and correlate with the end-user experience:
\begin{itemize}
    \item \textit{Semantic Accuracy}. Measures if certain key blendshapes are activated enough as expected for each peak expression. The expected blendshape activation for each peak expression is predefined by artists (\cref{sec:artistdata}). This is the \textbf{most important} metric reflecting the expressivity of the tracker.
    \item \textit{Neutralness}. Measures if the blendshape coefficients are below certain thresholds when the user stays neutral.
    \item \textit{Smoothness}. Measures if the blendshape activation curves are smooth, by measuring the mean of the second order derivatives.
    \item \textit{Eye Closure}. Measures if the avatar's eyes are fully closed when the user fully closes their eyes, including winking.
    \item \textit{Mouth Closure}. Measures if the avatar's mouth is fully closed when the user fully closes their mouth.
\end{itemize}

The metrics are first computed for each expression recording, then averaged across all the expressions for each subject, and finally averaged across all the subjects as the dataset-level metrics. This three-level (recording / subject / dataset) aggregation not only gives us an overall evaluation of the system, but also allows us to effectively locate certain problematic recordings or subjects to improve the system.

Note that the metrics do not rely on the pseudo ground truth. In fact, we can use the metrics to evaluate both the ML model and the pseudo ground truth, as shown in \cref{table:ft_metric}. This is important to our proposed Iterative Distillation training strategy (see \cref{subsec:training-framework}).

\begin{table}[t]
\caption{We evaluate the quantitative metrics for the pseudo GT, the initial ML model trained only with the pseudo GT, and the final ML model trained with the various strategies elaborated in \cref{subsec:training-framework}. While at a moderate trade-off on \textit{Neutralness} and \textit{Smoothness}, the final model significantly improves on \textit{Semantic Accuracy} and \textit{Mouth Closure}, which improves the overall end-user experience in practice.} \label{table:ft_metric}
{
\small
\centering
\begin{tabular}{cccccc}
& \multicolumn{1}{p{0.85cm}}{\centering \textit{Semantic \\ Accuracy}}
& \multicolumn{1}{p{0.85cm}}{\centering \textit{Neutral- ness}}
& \multicolumn{1}{p{0.85cm}}{\centering \textit{Smooth- ness}}
& \multicolumn{1}{p{0.85cm}}{\centering \textit{Eye \\ Closure}}
& \multicolumn{1}{p{0.85cm}}{\centering \textit{Mouth \\ Closure}} \\
\hline
Pseudo GT & 0.392 & \textbf{0.912} & \textbf{0.940} & 0.916 & 0.703 \\
Init. Model & 0.403 & 0.902 & 0.832 & \textbf{0.944} & 0.856 \\
Final Model & \textbf{0.700} & 0.774 & 0.868 & 0.936 & \textbf{0.905} \\
\end{tabular}
}
\end{table}



\subsection{Impact of Iterative Distillation}\label{sec:ablation_iterative}
As briefly illustrated in \cref{table:ft_metric}, the iterative distillation is important for model accuracy, \eg, improving the \textit{Semantic Accuracy} from 0.4 to 0.7. Here we study its impact more thoroughly from several aspects.

\begin{figure}
    \centering
    \includegraphics[width=0.9\linewidth]{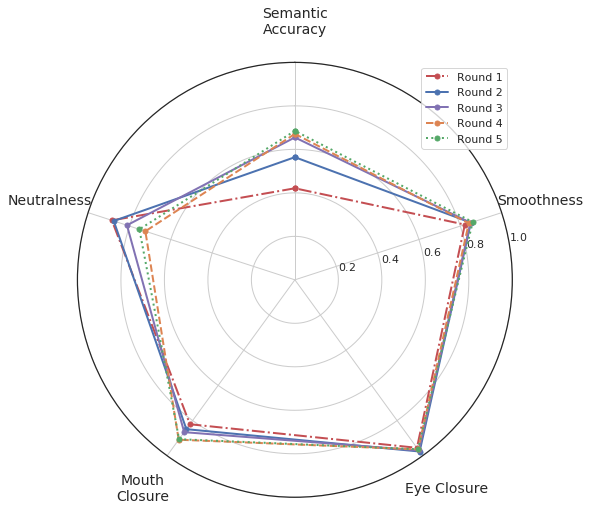}
    \caption{Quantitative metrics of the model trained at the end of each distillation iteration. The key metric, \textit{Semantic Accuracy}, keeps increasing till Round 5.}
    \label{fig:main_spider}
\end{figure}

\noindent\textbf{Iterations are necessary.} We first validate that multiple iterations of distillation are necessary to improve the model accuracy before having marginal gains. As demonstrated in \cref{fig:main_spider}, the \textit{Semantic Accuracy} has large improvements in the first two rounds and keeps increasing till Round 5.

\begin{figure}
    \centering
    \includegraphics[width=0.9\linewidth]{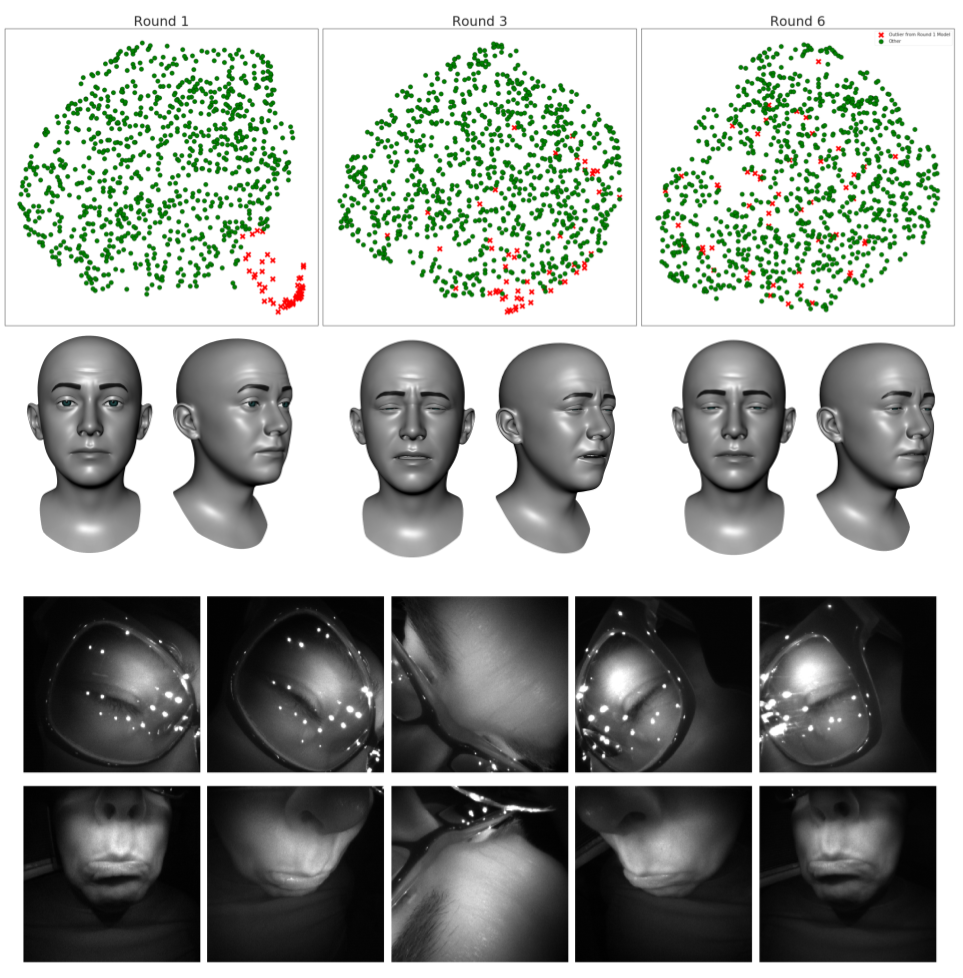}
    \caption{Iterative distillation can be viewed as denoising pseudo labels. The first row shows the model's latent feature space (with UMAP~\cite{SMG2020}) at different iterations. Red dots indicate outlier data points, among which we find an outlier sample and visualize its blendshape estimation in the second row. The bottom two rows show as reference the camera images of that outlier sample frame.}
    \label{fig:ensembling_help}
\end{figure}

\noindent\textbf{Distillation as label denoising.} The distillation process can be viewed as denoising the pseudo labels. We validate this by visualizing the ML model's latent feature space along the distillation iterations. As demonstrated in \cref{fig:ensembling_help}, there are some outliers in the Round 1 model's feature space which correspond to wrong blendshape coefficient estimation. As the distillation iteration goes, they gradually ``move'' into the distribution of inliers, and eventually have correct blendshape coefficient estimation.

\noindent\textbf{Model ensemble is important.} Lastly, we validate that model ensemble in the iterative distillation (\cref{alg:iterative_distillation}) is important. We conduct an ablation study that only one best performed model is selected in each distillation iteration, \ie, we train an initial model, using it as a teacher to train another model, and repeat. \cref{table:model_ensembling} shows the comparison between using model ensemble or not, where we can clearly see that using model ensemble performs much better on most of the metrics.

\begin{table}
\caption{Comparison between model ensemble or not in iterative distillation} \label{table:model_ensembling}
{
\small
\centering
\begin{tabular}{cccccc}
& \multicolumn{1}{p{1cm}}{\centering \textit{Semantic \\ Accuracy}}
& \multicolumn{1}{p{0.85cm}}{\centering \textit{Neutral- ness}}
& \multicolumn{1}{p{0.85cm}}{\centering \textit{Smooth- ness}}
& \multicolumn{1}{p{0.85cm}}{\centering \textit{Eye \\ Closure}}
& \multicolumn{1}{p{0.85cm}}{\centering \textit{Mouth \\ Closure}} \\
\hline
w/o Ensemble & 0.661 & 0.601 & 0.822 & 0.907 & \textbf{0.926} \\
w/ Ensemble & \textbf{0.700} & \textbf{0.774} & \textbf{0.868} & \textbf{0.936} & 0.905 \\
\end{tabular}
}
\end{table}

\subsection{Limitations}
At times, the resolution and placement of our IR cameras may restrict the level of detail that our system can capture. This is especially noticeable for users with obstructed facial features, like lips covered by facial hair or eyebrows hidden by thick glasses. Besides, although we aimed to make our system easily adoptable by implementing a blendshape model, this representation has certain inherent limitations. Since the blendshape bases are manually designed, they might not be the most optimal for achieving a compact representation and capturing subtle movements. Additionally, since the bases can have linear dependencies, multiple sets of blendshape weights can produce similar expressions leading to semantic ambiguities.

\section{Conclusion}
We have demonstrated the feasibility of achieving high-quality facial animation in real-time using a VR headset without the need of manual assistance, such as user calibration. Robust tracking is achieved by \Circle{1} embedding a set of IR cameras at strategic locations within the HMD, \Circle{2} collecting a rich and high-quality dataset of images and labels, and \Circle{3} developing a novel training framework to improve the accuracy of our ML model. Enhancing our ML model with audio and temporal information are future research areas that could help increase the realism even under extreme occlusions due to facial hairs or other obstructions.
We believe that our work will inspire further contributions in the development of consumer-level VR face trackers and will pave the way for new interaction metaphors and social presence experiences.
{
    \small
    \bibliographystyle{ieeenat_fullname}
    \bibliography{main}
}


\end{document}